\documentclass[conference]{IEEEtran}
\IEEEoverridecommandlockouts
\usepackage{cite}
\usepackage{amsmath,amssymb,amsfonts}
\usepackage{algorithmic}
\usepackage{graphicx}
\usepackage{textcomp}
\usepackage{dblfloatfix} 
\usepackage{amsmath, amssymb, mathtools}
\usepackage{xcolor}
\usepackage{hyperref}
\usepackage{booktabs}
\usepackage{makecell}

\def\BibTeX{{\rm B\kern-.05em{\sc i\kern-.025em b}\kern-.08em
    T\kern-.1667em\lower.7ex\hbox{E}\kern-.125emX}}
\begin{document}

\title{One-Step Diffusion for Perceptual Image Compression\\}

\author{Yiwen Jia, Hao Wei, Yanhui Zhou, Chenyang Ge}

\author{
Yiwen Jia$^1$, Hao Wei$^1$, Yanhui Zhou$^2$ and Chenyang Ge$^{1,*}$ \\
$^1$Institute of Artificial Intelligence and Robotics, Xi’an Jiaotong University, China \\
$^2$School of Information and telecommunication, Xi’an Jiaotong University, China \\
jiayiwen@stu.xjtu.edu.cn, haowei@stu.xjtu.edu.cn, zhouyh@mail.xjtu.edu.cn, cyge@mail.xjtu.edu.cn
}
\maketitle

\begin{abstract}
Diffusion-based image compression methods have achieved notable progress, delivering high perceptual quality at low bitrates. However, their practical deployment is hindered by significant inference latency and heavy computational overhead, primarily due to the large number of denoising steps required during decoding. 
To address this problem, we propose a diffusion-based image compression method that requires only a single-step diffusion process, significantly improving inference speed. 
To enhance the perceptual quality of reconstructed images, we introduce a discriminator that operates on compact feature representations instead of raw pixels, leveraging the fact that features better capture high-level texture and structural details.
Experimental results show that our method delivers comparable compression performance while offering a 46$\times$ faster inference speed compared to recent diffusion-based approaches. The source code and models are available at \href{https://github.com/cheesejiang/OSDiff}{\textit{https://github.com/cheesejiang/OSDiff}}.

\end{abstract}

\begin{IEEEkeywords}
Image compression, diffusion models, one step sampling.
\end{IEEEkeywords}

\section{Introduction}
\begin{figure}[htbp]
  \centering
  \includegraphics[width=0.48\textwidth]{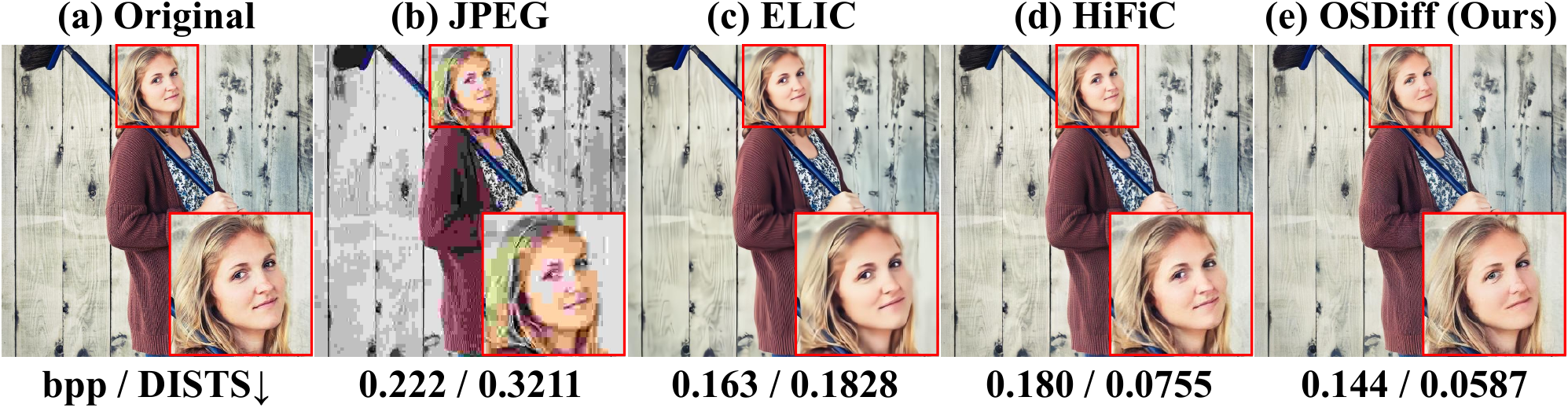}\vspace{-0.3cm}
  \caption{Qualitative comparisons of different methods on test datasets.}
  \label{intro}
\end{figure}

With the growing demand for digital images, efficient image compression has emerged as essential. Traditional methods like JPEG\cite{wallace1991jpeg} rely on hand-crafted heuristics, struggle to handle diverse content, and often introduce visible artifacts, as shown in Fig.\ref{intro}(b). 
On the other hand, learned image compression methods\cite{balle2016end, joint, cheng2020learned ,he2022elic, zhu2022transformer, liu2023learned, xie2021enhanced}, which aim to optimize the rate-distortion trade-off \cite{shannon1959coding}, have gained popularity as effective alternatives to traditional approaches due to their superior compression performance. 
However, they usually produce over-smooth results, especially at low bitrates (see Fig.\ref{intro}(c)).

To overcome these limitations, perceptual-driven generative image compression methods \cite{p-d, r-d-p, deep, universal, EIR, yang2023lossy} based on generative adversarial networks (GAN)
\cite{agustsson2019generative, compressnet, yan2021perceptual, he2022po, agustsson2023multi} and diffusion models \cite{careil2023towards, li2024towards} have been proposed. However, GAN-based methods experience significant performance degradation in extremely low bitrate scenarios. For example, HiFiC \cite{hific} generates reconstructed results with unrealistic details at low bitrates (Fig.\ref{intro}(d)).
By contrast, diffusion models have shown great potential for image compression \cite{hoogeboom2023high, li2024towards, relic2024lossy, lei2023text+}, leveraging their powerful generative capabilities \cite{ho2020denoising}.
Although these methods produce realistic reconstructions, a major flaw remains: \textit{The denoising process in diffusion models typically involves numerous iterative steps, leading to significant inference latency and computational overhead}.

In this paper, we propose a diffusion-based perceptual image compression, named \textbf{OSDiff}, that enables decoding in one denoising step.
Specifically, our method leverages the generative prior embedded in the pre-trained Stable Diffusion \cite{rombach2022high}, facilitating more realistic image reconstructions (Fig.\ref{intro}(e)). Unlike previous diffusion-based methods \cite{li2024towards} using 50 steps, our approach accelerates the denoising process in a single step. Specifically, the denoising process starts from the noisy image, and the clean image is generated in a single sampling step. This substantially accelerates the inference process and reduces computational cost.
To further improve the perceptual quality of reconstructed results, we introduce a discriminator that operates in the latent feature domain to distinguish between generated and original images while avoiding incurring additional computational overhead for image encoding or decoding during inference.

In summary, our main contributions are as follows:
\begin{itemize}
    \item We propose a diffusion-based perceptual image compression approach that performs one-step diffusion, significantly reducing inference latency and computational cost.
    \item We introduce a discriminator that operates in a designated feature space to further enhance the perceptual quality of reconstructed images.
\end{itemize}

\begin{figure*}[htbp]
  \centering
  \includegraphics[width=0.73\textwidth]{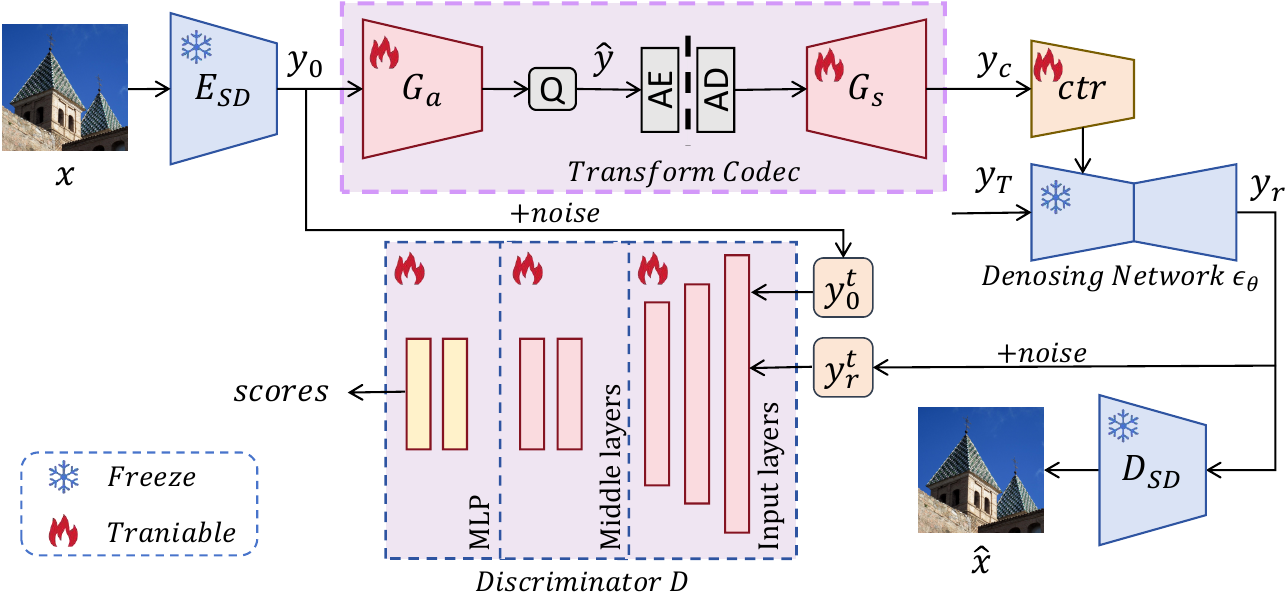}\vspace{-0.3cm}
  \caption{The framework of our \textbf{OSDiff}, which is composed of the frozen encoder–decoder pair ($E_{SD}$, $D_{SD}$), the modules $G_a$ and $G_s$, the discriminator $D$, and the denoising network $\epsilon_{\theta}$ with control module ($ctr$ for short). During training, only the modules $G_a$, $G_s$, $ctr$, and the discriminator $D$ are optimized through the loss function, while the parameters of other components remain frozen.}
  \label{framework}
\end{figure*}

\section{Method}

First, we present the overall framework. Then we adopt a one-step sampling strategy to accelerate the denoising process and reduce computational overhead during inference. Finally, we devise a discriminator to align the distribution of reconstructed images with that of the ground truth images. 
\subsection{Overview}\label{3_1}
As shown in Fig.\ref{framework}, the input $x$ is first encoded by the stable diffusion's encoder $E_{SD}$ to obtain $y_0$ in the $4\times$ downsampled latent space. Then encoder $G_a$ maps $y_0$ to the $8\times$ downsampled space. The latent features are then quantized into $\hat{y}$ that are losslessly compressed using arithmetic coder. The decompressed features $\hat{y}$ are upsampled once through the decoder $G_s$ to obtain $y_{c}$. Then the denoising network $\epsilon_{\theta}$ with control module \cite{ctr}, which shares the same encoder and middle blocks as $\epsilon_{\theta}$, takes $y_c$ as the condition to reconstruct realistic features $y_r$ from noisy $y_T$. Finally, the reconstructed features $y_{r}$ are decoded by $D_{SD}$ to get the reconstructed images. The discriminator $D$ is used to minimize the distributional gap between the reconstructed images generated by OSDiff and the ground truth images. The entire process can be formulated as follows, where $y_0^t$, $y_r^t$ denote the noisy version of $y_0$, $y_r$ after $t$ steps of the forward diffusion process:
\begin{flalign}
&\hspace{2em} y_0 = E_{SD}(x), \quad \hat{y} = Q(G_{a}(y_0)), \quad y_c = G_{s}(\hat{y}), & \label{eq:encoding} \\
&\hspace{2em} y_c,y_T \stackrel{\epsilon_\theta}{\longrightarrow} y_r, & \label{eq:denoising} \\
&\hspace{2em} \hat{x} = D_{SD}(y_r), & \label{eq:decoding}\\
&\hspace{2em} scores = D(y_0^t,y_r^t). & 
\label{eq:discrim}
\end{flalign}

\subsection{One-Step Sampling Diffusion}
To accelerate the denoising process, we adopt the one-step sampling strategy to accelerate the diffusion process.
Diffusion models consist of a forward process and a reverse process. In the forward process, the clean image $x_0$ is gradually corrupted by adding predefined Gaussian noise. When $T{=}1000$, the resulting image $x_T$ becomes nearly indistinguishable from pure noise. In stable diffusion, this process is performed in the latent space. The forward process can be expressed as follows:
\begin{equation}
q(y_t\mid y_0) = \mathcal{N} \left( y_t; \, \sqrt{\bar{\alpha}_t} \, y_0, \, (1 - \bar{\alpha}_t) \, \epsilon \right),
\end{equation}
where $\epsilon \sim \mathcal{N}(0, \mathbf{I}),\;\alpha_t = 1 - \beta_t,  \text{and}\;\bar{\alpha}_t = \prod_{i=1}^t \alpha_i$. $\beta_t \in (0,1)$ controls the noise level.
In the reverse process, the denoising network estimates the injected noise and performs multi-step denoising to progressively recover the clean image feature. We  can use network $\epsilon_\theta$ to predict the noise $\hat\epsilon = \epsilon_\theta({y}_t, {y}_c, t)$. The multi-step denoising process can be expressed as follows:
\begin{align}
&q_\theta({y}_{t-1} \mid {y}_t) 
= \notag \\
\mathcal{N}\Bigg(
&\frac{1}{\sqrt{\alpha_t}} \left(
{y}_t - \frac{1 - \alpha_t}{\sqrt{1 - \bar{\alpha}_t}} \epsilon_\theta(y_t, y_c, t)
\right),
\frac{1 - \bar{\alpha}_{t-1}}{1 - \bar{\alpha}_t} \beta_t \epsilon
\Bigg).
\end{align}
Then the objective function can be written as follows:
\begin{equation}
\mathcal{L}_{\text{Diff}} = \mathbb{E}_{y_0, t, \epsilon} \left\| \epsilon - \epsilon_\theta(y_t, y_c, t) \right\|^2.
\end{equation}
The above describes the multi-step sampling process, which requires substantial time and computational resources. In this paper, we propose a one-step sampling strategy that directly samples the clean image feature $\hat{y}_0$ from the noisy image feature $y_t$ in one step, where $\hat{y}_0$ exactly corresponds to the reconstructed feature ${y}_r$. The formulation is as follows:
\begin{equation}
\hat{y}_0 = \frac{y_t - \sqrt{1 - \bar{\alpha}_t} \hat\epsilon}{\sqrt{\bar\alpha}_t}.
\end{equation}

\begin{figure}[htbp]
  \centering
  \includegraphics[width=0.30\textwidth]{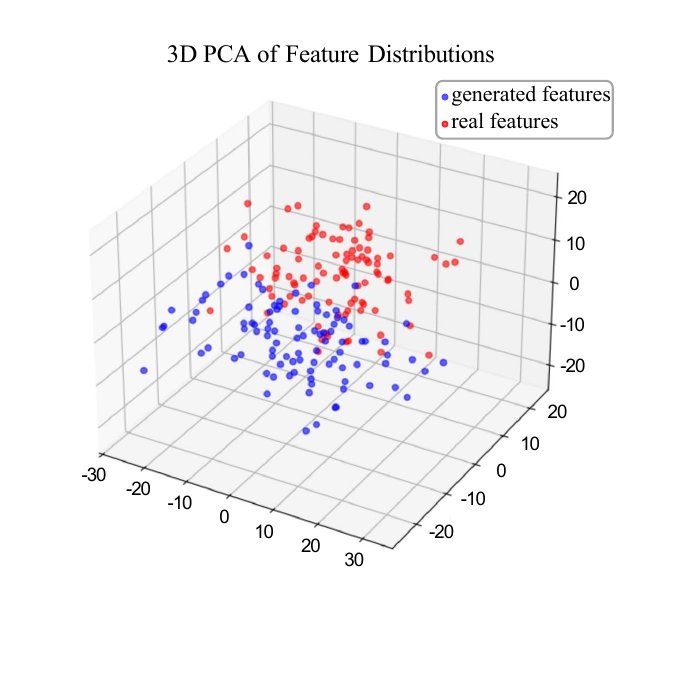}\vspace{-0.3cm}
  \caption{The distribution of generated features and real features in the specific lantent feature space.}
  \label{dis}
\end{figure}
\subsection{Discriminator in Stable Diffusion space}

%
To enhance the realism of generated images, a discriminator is introduced. Unlike previous methods act in the pixel space \cite{lightweight}, we propose to discriminate in the latent feature space.
%
As Fig.~\ref{dis} shows, there exists a distribution gap between the intermediate data of generated and real features in the middle layers of the U-Net. Based on this observation, we constrain the generated image in this feature space to better resemble the original image.

Specifically, for the discriminator, we obtain the $y_0^t$ and $y_r^t$ through the forward process from $y_0$ and $y_r$, respectively. The features $y_0$, derived from encoding the original image $x_0$ with the VAE encoder, serve as the ground truth. The features $y_r$ are the reconstructed features derived from the one-step sampling strategy. The input and middle layers of the U-Net process the features $y_0$ and $y_r$, which represent the real and generated images, respectively, to extract intermediate features. Then, these features are passed to a MLP for discrimination scoring.


\subsection{Model Objectives}
For the one-step sampling generator, the target losses are as follows:

\subsubsection{Diffusion Loss}
We use the clean features $y_r$ and the target features to calculate the diffusion loss to optimize the denoising network:
\begin{equation}
\mathcal{L}_{\text{diff}} = \left\lVert y_0 - y_r \right\rVert^2.
\end{equation}

\subsubsection{Rate Loss}
This loss serves to optimize the rate performance:
\begin{equation}
\mathcal{L}_{\text{rate}} = R(\hat{y}).
\end{equation}

\subsubsection{Latent Feature Loss}
This loss is used to optimize the codec in the transform process:
\begin{equation}
\mathcal{L}_{\text{feature}} = \left\lVert y_0 - y_c \right\rVert^2.
\end{equation}


\subsubsection{Generator Loss}
This loss is the generator loss of GAN, where $y_r^t$ denotes the noisy version of $y_r$ after $t$ steps of the forward diffusion process:
\begin{equation}
\mathcal{L}_G = -\mathbb{E}_{t} \left[ \log D(y_r^t) \right].
\end{equation}
In total, the generator loss is defined as:
\begin{equation}
\mathcal{L}_{G_{total}} =\lambda_1\mathcal{L}_{\text{diff}} + \lambda_2\mathcal{L}_{\text{rate}} + \lambda_3\mathcal{L}_{\text{feature}} + 
\lambda_4\mathcal{L}_G.
\end{equation}
To optimize the proposed discriminator, we use:
\begin{equation}
\begin{split}
\mathcal{L}_{D_{total}} &= -\mathbb{E}_{t} \left[ \log \left(1 - D(y_r^t)\right) \right] \\
&\quad - \mathbb{E}_{t} \left[ \log D(y_0^t) \right],
\end{split}
\end{equation}
where $y_0^t$ denotes the noisy version of $y_0$ after $t$ steps of the forward diffusion process.

\section{Experiments}

\begin{figure*}[htbp]
  \centering
  \includegraphics[width=1.0\textwidth]{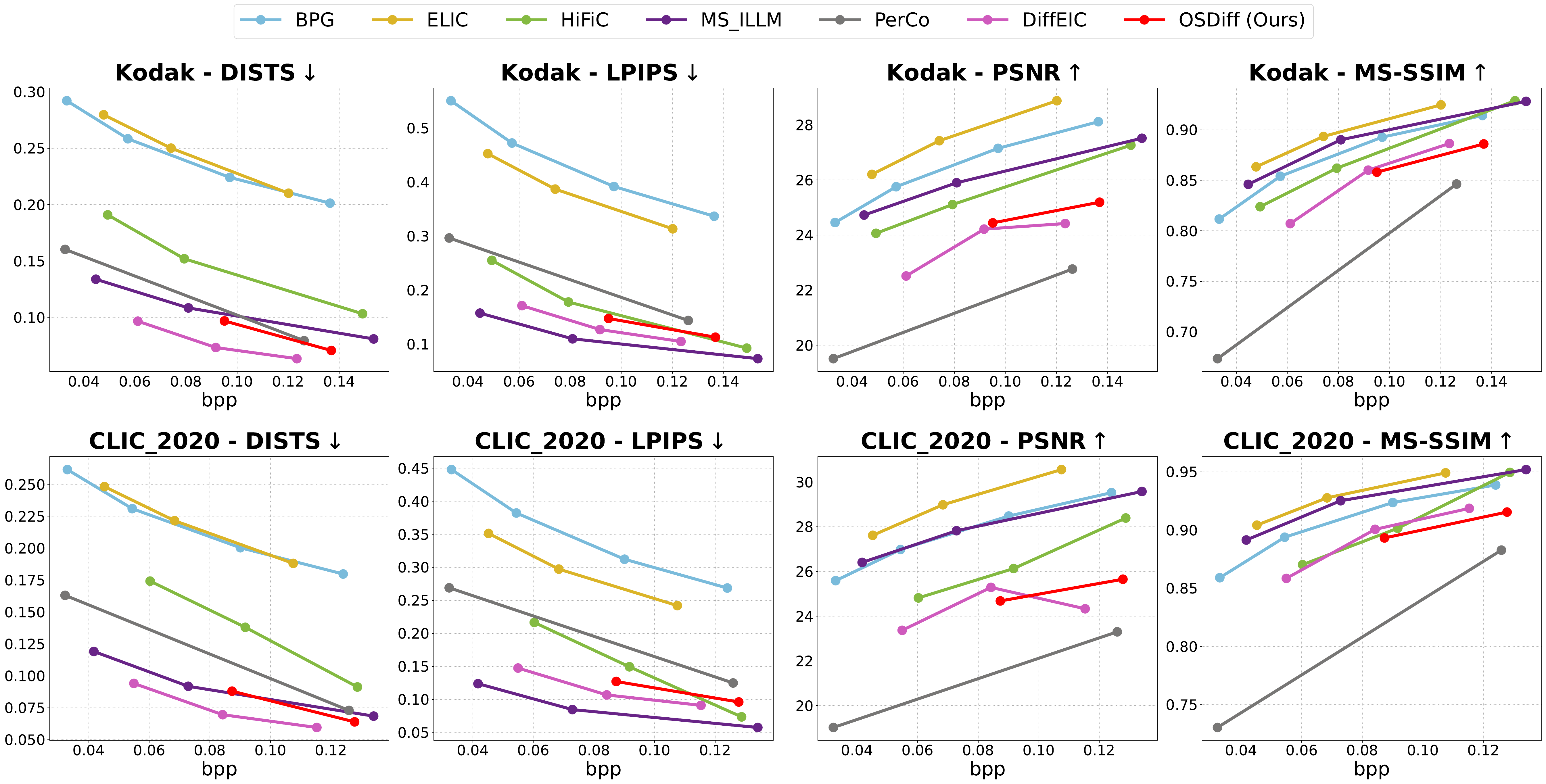}\vspace{-0.3cm}
  \caption{Quantitative comparisons with the state-of-the-art method on test datasets.}
  \label{BDcurve}
\end{figure*}

\subsection{Experimental Setup}
\label{sec:4.1}
\subsubsection{Datasets}
We train OSDiff on the training split of \textbf{LSIDR}~\cite{lsidr_dataset} dataset, which consists of 84{,}991 high-quality images. All images are randomly cropped to a resolution of 512$\times$512 pixels for input. For evaluation, we adopt several commonly used image compression benchmark datasets: 
\textbf{Kodak}\cite{kodak}, which contains 24 images with a resolution of 512$\times$768 pixels; 
the test set of \textbf{CLIC\_2020}\cite{2020clic}, consisting of 428 test images with 2K resolution. For the CLIC\_2020 dataset, each image is first resized proportionally such that the shorter side is 768 pixels, followed by a center crop to obtain a final resolution of 768$\times$768 pixels.
\subsubsection{Metrics}
In this work, we aim to optimize the trade-off among rate, distortion, and perceptual quality. We evaluate performance using both distortion (PSNR, MS-SSIM\cite{ssim}) and perceptual metrics (LPIPS\cite{lpips}, DISTS\cite{dists}). 
\subsubsection{Training Details}
We load the denoising network using the pretrained weights from Stable Diffusion 2.1. To support the discriminator, which operates in the feature space, we extract features using a subnetwork composed of the input and middle layers of the pre-trained UNet, denoted as $f_d$. Both the parameters of $f_d$ and the discriminator are updated during training. To reduce GPU memory usage, we adopt the AdamW8bit\cite{8bit} optimizer with parameters $\beta_1 = 0.9$ and $\beta_2 = 0.999$, and set the learning rate at $1 \times 10^{-4}$. We set $\lambda_1$, $\lambda_3$ and $\lambda_4$ to 1, 2 and 0.01, respectively, and choose $\lambda_2$ from $\{$1, 2$\}$ to achieve different coding bitrates. All experiments are performed on a single NVIDIA GeForce RTX 4090 GPU.

\subsection{Methods Comparisons}
\label{sec:4.2}
We compare our method with traditional, learning-based and diffusion-based image compression methods, including
\textbf{BPG}~\cite{bpg},
\textbf{ELIC}~\cite{he2022elic},
\textbf{HiFiC}~\cite{hific},
\textbf{MS-ILLM}~\cite{illm},
\textbf{PerCo}~\cite{careil2023towards}, and
\textbf{DiffEIC}~\cite{li2024towards}.
\subsubsection{Qualitative Comparisons}
Fig.\ref{BDcurve} shows the rate-distortion-perception curves of different methods at low bitrates. 
The following conclusions can be drawn: \emph{i)} Compared to PerCo, OSDiff achieves better perceptual and distortion quality although using one-step sampling; \emph{ii)} Compared to DiffEIC with 50-step sampling, OSDiff exhibits a slight drop in perceptual quality due to the one-step sampling strategy. But its distortion metric (PSNR) surpasses that of DiffEIC and does not degrade as the bitrate increases($>0.1bpp$); \emph{iii)} Overall, OSDiff achieves the best performance on the DISTS metric compared to other non-diffusion-based methods.

\subsubsection{Quantitative Comparisons}
\begin{figure}[htbp]
  \centering
  \includegraphics[width=0.49\textwidth]{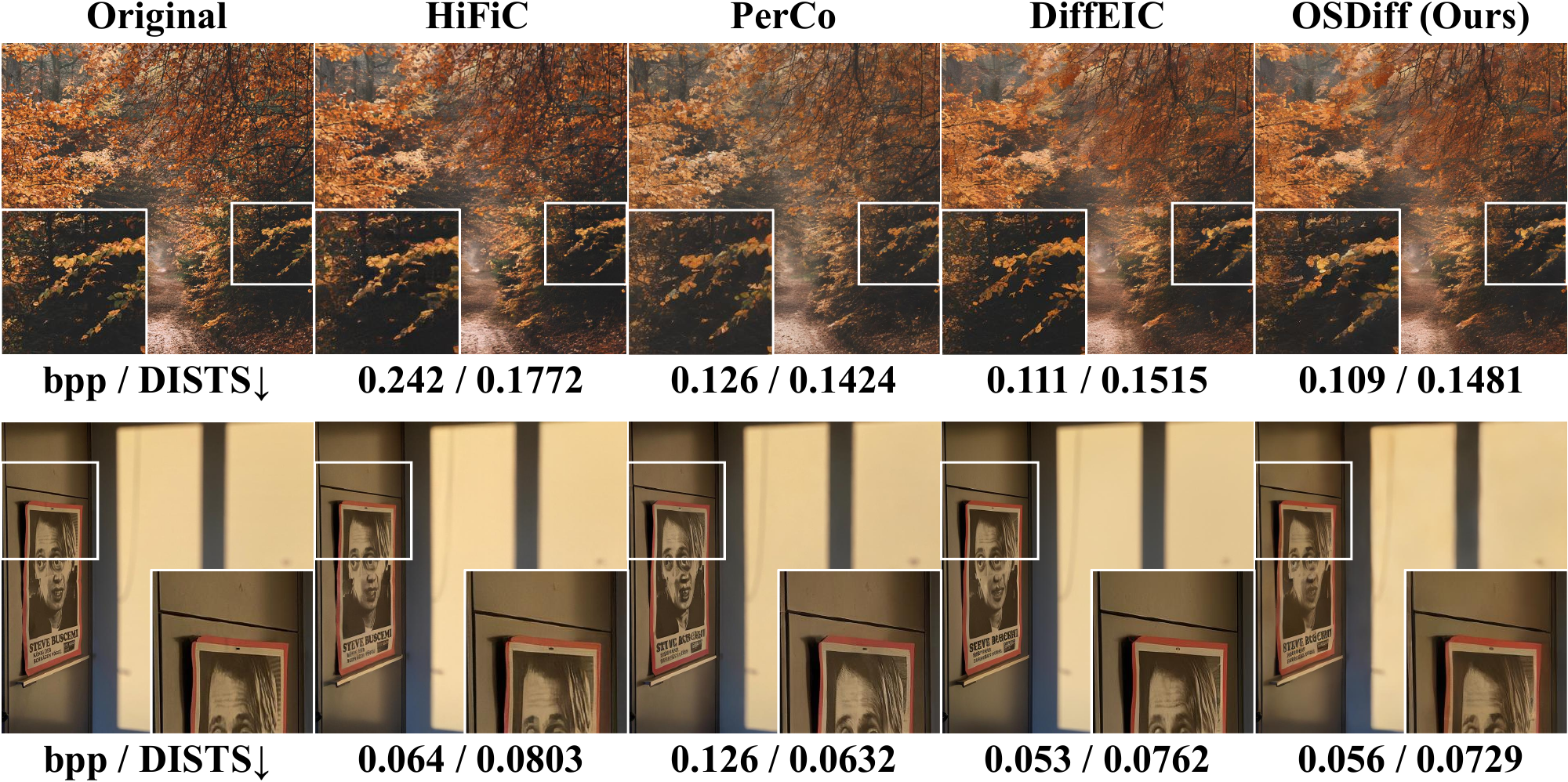}\vspace{-0.3cm}
  \caption{Qualitative comparisons of different methods on test datasets.}
  \label{imagecompare}
\end{figure}

We provide visual results in Fig. \ref{imagecompare}. Compared to HiFiC and PerCo, OSDiff achieves superior visual quality. Moreover, OSDiff achieves perceptual quality that is close to that of DiffEIC, while significantly accelerating the inference process and reducing computational cost. For example, OSDiff more faithfully preserves the leaf structure (especially the green leaf), the small holes on the wall, and the striped background behind the man's hair.

\subsubsection{Inference latency}
We compare the inference latency of three diffusion-based methods. For the PerCo method, we directly report the result from the original paper. As Table \ref{latency} shows, for an image of size 512$\times$768, OSDiff achieves a decoding time of only 0.060 seconds on an RTX 4090, which is approximately 50 times faster than DiffEIC. This reflects our original design intention of reducing inference latency.

\begin{figure}[htbp]
  \centering
  \includegraphics[width=0.5\textwidth]{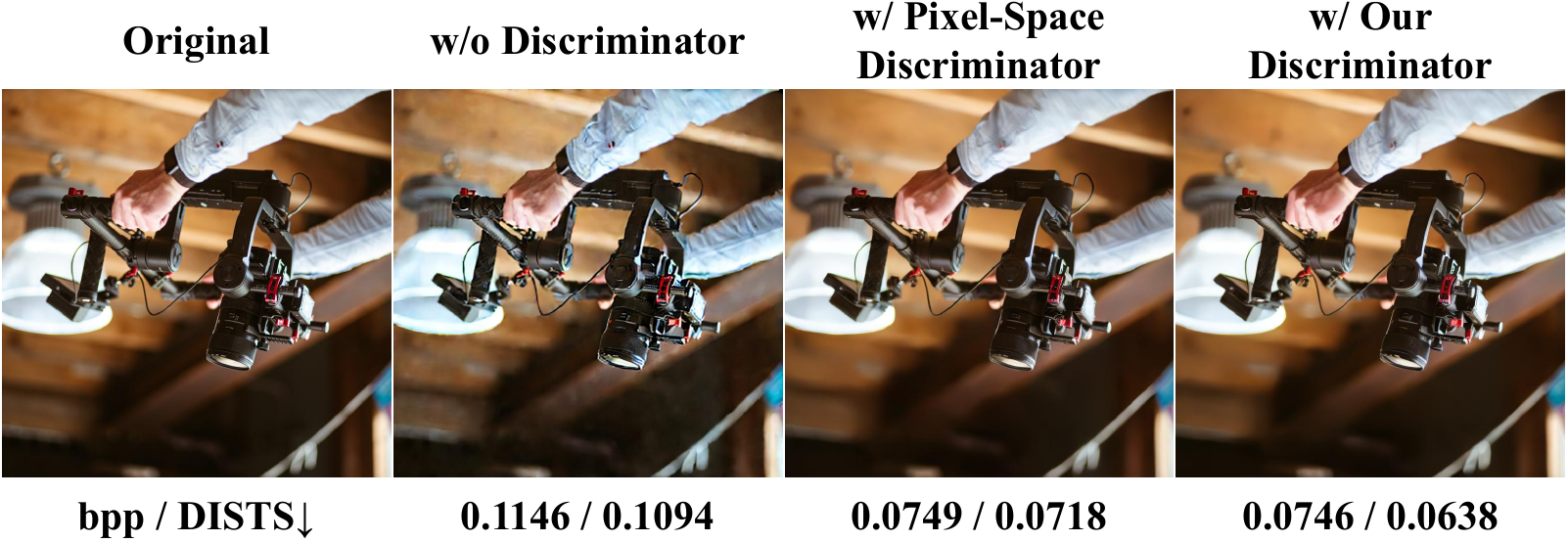}\vspace{-0.3cm}
  \caption{Visual results for validating the effectiveness of the discriminator D.}
  \label{imageablation}
\end{figure}

\begin{table}[htbp]
\setlength{\tabcolsep}{4pt}  
\centering
\small
\caption{Inference latency comparison of diffusion-based methods for a $512\times 768$ image.}
\label{latency}
\begin{tabular}{lc|cccc}
\toprule
\textbf{Method} & 
\makecell{\textbf{Sampling}\\\textbf{Steps}} & 
\makecell{\textbf{Encoding}\\\textbf{Time (s)}} & 
\makecell{\textbf{Decoding}\\\textbf{Time (s)}} & 
\textbf{Device} \\
\midrule
PerCo  & 5  & 0.080 & 0.665 & A100 \\
PerCo & 20 & 0.080 & 2.551 & A100 \\
DiffEIC & 50 & 0.093 & 2.761  & RTX 4090 \\
\textbf{OSDiff (Ours)} & \textbf{1} & 0.101 & \textbf{0.060} & RTX 4090 \\
\bottomrule
\end{tabular}
\end{table}

\subsection{Ablation Experiments}
\label{sec:4.3}
As Table \ref{ablation} shows, we conducted experiments with three different settings: removing the discriminator, using a pixel-space discriminator, and using our discriminator. It is evident that introducing a discriminator in the latent space significantly improves the distortion and perception quality of the reconstructed image compared to the setting without a discriminator. Additionally, our discriminator outperforms the pixel-space discriminator in terms of both distortion and perception quality, further demonstrating its effectiveness. The visualization results are shown in Fig.\ref{imageablation}. The visual quality of the generated images is significantly improved after introducing our discriminator, with richer details and better alignment with human perception.

\renewcommand{\arraystretch}{0.9}  
\begin{table}[htbp]
  \centering
  \caption{BD-rate computed on the CLIC\_2020 dataset using MS-SSIM for distortion and DISTS for perception. Demonstrates the effectiveness of the proposed discriminator D.}
  \begin{tabular}{c|ccc}
    \toprule
    \textbf{Methods} & \multicolumn{3}{c}{\textbf{BD-rate (\%)}} \\
    \cmidrule(lr){2-4}
     & \textbf{Distortion} & \textbf{Perception} & \textbf{Average} \\
    \midrule
    w/o Discriminator & 0 & 0 & 0 \\
    \midrule
    w/ Pixel-Space Discriminator &  -69.99\%  & -80.87\%  & -75.43\% \\
    \midrule
    w/ \textbf{Our Discriminator} &  \textbf{-72.60\%}  &  \textbf{-82.09\%}  & \textbf{-77.35\%} \\
    \bottomrule
  \end{tabular}
  \label{ablation}
\end{table}

\section{Conclusion}
In this work, we propose a one-step diffusion-based image compression method which significantly reduces inference latency and computational complexity during the inference compared to existing diffusion-based approaches. To further enhance reconstruction quality, we introduce a discriminator operating in a designated feature space. Experimental results demonstrate that our method achieves comparable or even superior performance to existing methods in perceptual metrics while being substantially more efficient.

\clearpage

\bibliographystyle{ieeetr}  
\bibliography{main}  

\end{document}